\documentclass[letterpaper,11pt,leqno]{article}
\usepackage{paper}
\usepackage{amsmath}
\usepackage{enumitem}
\newcommand{\RomanNumeralCaps}[1]
    {\MakeUppercase{\romannumeral #1}}

\begin{document}

\title{Homeostatic Continual Learning}
\author{Yue Jin\\
Nokia Bell Labs France\\
yue.1.jin@nokia-bell-labs.com}

\maketitle

I formulate a Continual Learning problem for an autonomous agent that imitates the one animals face. The agent has intents to satisfy. It experiences contexts where different combinations of features are active and impact intent outcomes. I propose a method named ``Homeostatic Continual Learning'' that imitates the one animals use. The agent learns incrementally environment data and features relevant for their intents from the impact of contexts, gradually completes its model and policy and performs well in more and more contexts. The core of the method is to find outliers in the environment data when the agent experiences an outlier in its intent outcome after the context changes. This method enables the agent to learn continuously in changing contexts without catastrophic forgetting. When pursuing many intents concurrently, the agent may build a transformer-like world model with the method where it factorizes the objects in the world into features, abstract objects into comparable instances of concepts and map concepts to intents through features. The world model is composed of semantic nodes and explainable by nature. I discuss the works needed to render the method practical, the connections to many fields in Artificial Intelligence and the broader implications of the method.

\section{Introduction}

In Pavlov's experiment, Pavlov presented a stimulus (e.g. the sound of a bell) and then gave the dog food; after a few repetitions, the dogs started to salivate in response to the stimulus \cite{pavlov1938lectures}. The dog learns that the stimulus is relevant for getting food. Through evolution hundreds of millions of years, animal brains develop the ability to continuously adapt to changing environmental conditions and to come up with new solutions on the fly \cite{durstewitz2025neuroscience}. These brains don't suffer from catastrophic forgetting and maintain the performance in previous tasks after learning new tasks. The state-of-the-art autonomous agents are yet to acquire these capabilities. 

In this paper, I formulate a Continual Learning problem for an autonomous agent that imitates the one animals face. The agent has intents to satisfy. It experiences contexts where different combinations of features are active and impact intent outcomes. I propose a method named ``Homeostatic Continual Learning'' that imitates the one animals use. The agent learns incrementally environment data and features relevant for their intents from the impact of contexts. The core of the method is to find outliers in the environment data when the agent experiences an outlier in its intent outcome. After the context changes, the agent observes that the model and policy it learns in previous contexts stop working. It detects a new context from this outlier in its outcome. It identifies a subset in the environment data that have outliers, defines a new feature over the subset and expands its model and policy with the subset. Over time and many iterations the agent gradually completes its model and policy and performs well in more and more contexts. This method enables the agent to learn continuously in changing contexts without catastrophic forgetting. 

When pursuing many intents concurrently, the agent may build a world model with the method where it factorizes the objects in the world into features, abstract objects into comparable instances of concepts and map concepts to intents through features. The learning results is a transformer-like world model that is composed of semantic nodes and explainable by nature. The factorization enables the agent to reduce the complexity of its decision and create new combinations of features to better adapt to new contexts. 

The paper is organized as the following. In Section \ref{sec:problem}, I formulate the Continual Learning problem. In Section \ref{sec:example}, I use an example to illustrate the problem and the method. In Section \ref{sec:method}, I explain the method and some of its extensions. In Section \ref{sec:findings}, I present merits of the method, including solving the catastrophic learning problem and building the semantic and explainable world model. In Section \ref{sec:works}, I discuss the works needed to render the method practical. In Section \ref{sec:connections}, I discuss the connections of the method with many fields in Artificial Intelligence. I conclude in Section \ref{sec:implications} with a discussion on broader implications of the method. 

\section{The Problem} \label{sec:problem}

\begin{figure}[t]
\centering
\includegraphics[width=0.8\columnwidth]{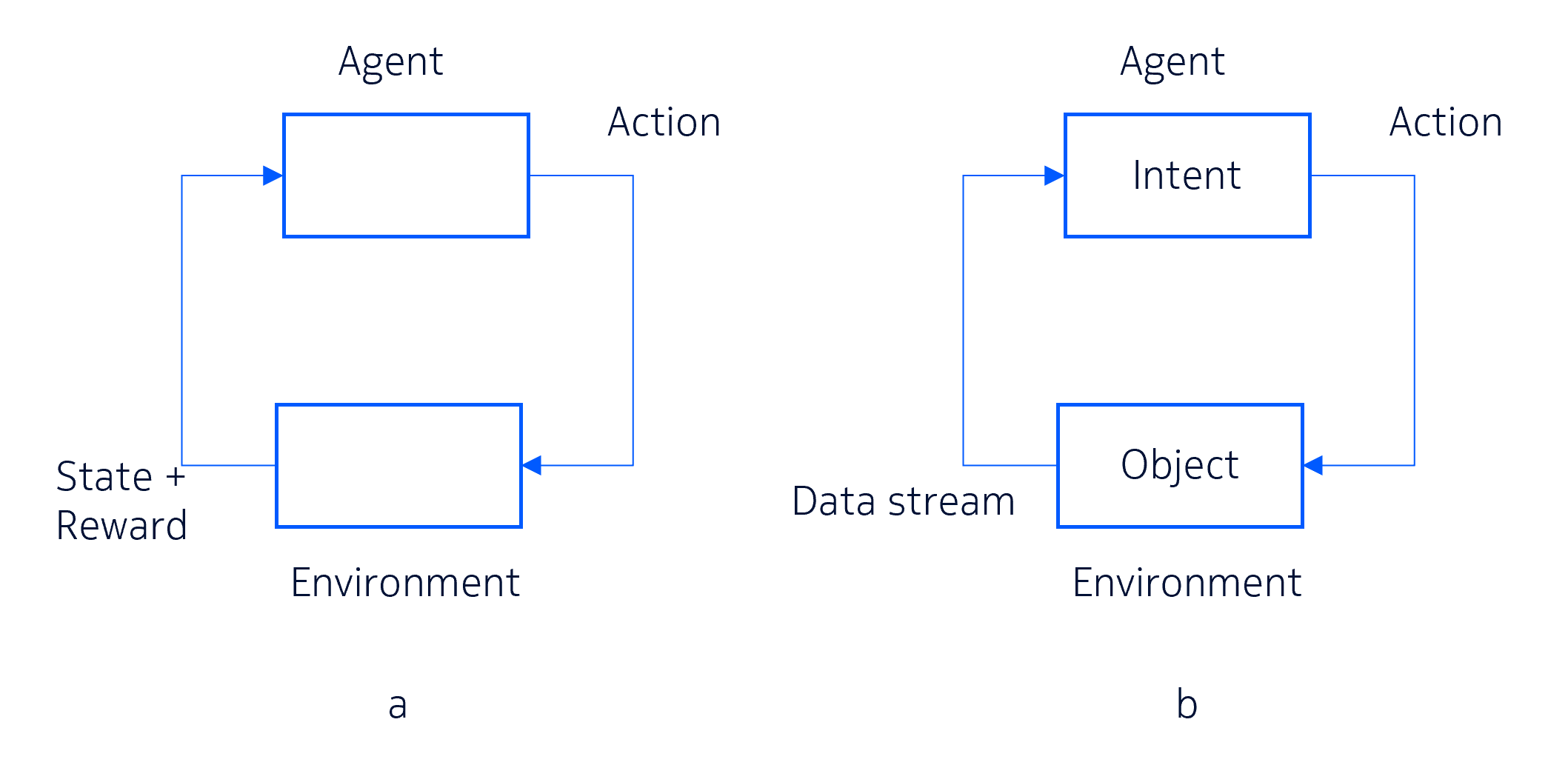}
\caption{The problem: the agent needs to identify the data streams relevant for its intent. a) Reinforcement Learning problem. b) Continual Learning problem in this work that abstracts the problem more realistically.}
\label{fig:problem0}
\end{figure}

Conventionally we abstract the learning problem animals face as a Reinforcement Learning problem \cite{sutton1998reinforcement} and formulate the problem for an autonomous agent analogously. As depicted in Figure \ref{fig:problem0}a, the agent receives states and rewards from its environment. It finds the optimal policy to maximize a cumulative reward. But one issue is rarely discussed and remains unsolved. That is how the agent can learn what data to include in the states from the vast variety of data in its environment. The problem depicted in Figure \ref{fig:problem0}b abstracts more realistically the problem animals face. I define the problem as follows. 

An agent needs to satisfy an intent. An intent is a desire to maintain a specific condition at a certain value. Real world examples include ``not go hungry'' or ``satisfy the service level agreement (SLA)''. The agent needs to find the right actions for different situations, i.e. policies, to satisfy the intent. 

It receives many data streams from the objects in its environment. Each object has multiple features. For instance, an apple has shape, color, taste, size, etc. These features generate data streams. For instance, color and shape can be represented in pixels. In general, a single feature can generate one or multiple data streams and a data stream can be generated by the interactions of multiple features. As an example for the latter, the color of an apple is generated by the apple and the lighting condition. The agent isn't given the features and needs to define them from data streams. Not all the features are relevant to the intent. Consequently not all the data steams are. The agent needs to identify the relevant data streams, define the relevant features, build a model that maps the data streams to the intent outcome and find the policies. Both the intent and the data streams are persistent. 

The data don't show the effects of all the relevant features at once. Some features stay constant or are masked for extended period of time and provide no information on its effect. I refer to a feature that doesn't show its effect as a ``static'' feature and one that does as an ``active'' feature. For instance, a mouse needs to press a button of specific shape and color to get food; if it is shown multiple buttons of the same shape but different colors, the shape is a static feature and the color is an active feature. Different combinations of static and active features and their values form different contexts. A static feature becomes active when it is active in another context or when it changes its value from one context to another. For instance, the button shape becomes an active feature when it changes from circle to square from one context to another.  From the data streams in the contexts the agent experiences, it incrementally identifies the relevant data streams, defines the relevant features, builds a model and finds the policies. The model for the intent is incomplete before the agent identifies all the relevant features. Since the agent may never experience all the relevant features, the model may remain incomplete forever for some intents. This continual learning problem extends the Domain-Incremental Learning in \cite{van2022three}. 

I assume that each context last long enough for the agent to learn its model and policy in the context and the relationships between the data streams and the intent outcome are deterministic. In this work, I study the special case with the following assumptions.
\vspace{\baselineskip}
\begin{enumerate}[label=\textnormal{(\Roman*)}]
\item There is a bijection mapping between data streams and features. 
\item A static feature stays constant in all the contexts. It only becomes active by changing values from one context to another.
\item From one context to another, only 1 static feature changes values. The active features have the same set of values.
\end{enumerate}
\vspace{\baselineskip}
I'll discuss the relaxation of the assumptions in Section \ref{sec:method}.

I formulate the continual learning problem as follows. The agent has an intent to maintain a condition $y$ to a certain value $r$. It has a set $U$ from which it can select an action $u \in U$. After it takes the action $u$, it observes an outcome $y$ of the intent. It needs to find the action $u$ that satisfies the intent $y=r$.

\begin{figure}[t]
\centering
\includegraphics[width=0.8\columnwidth]{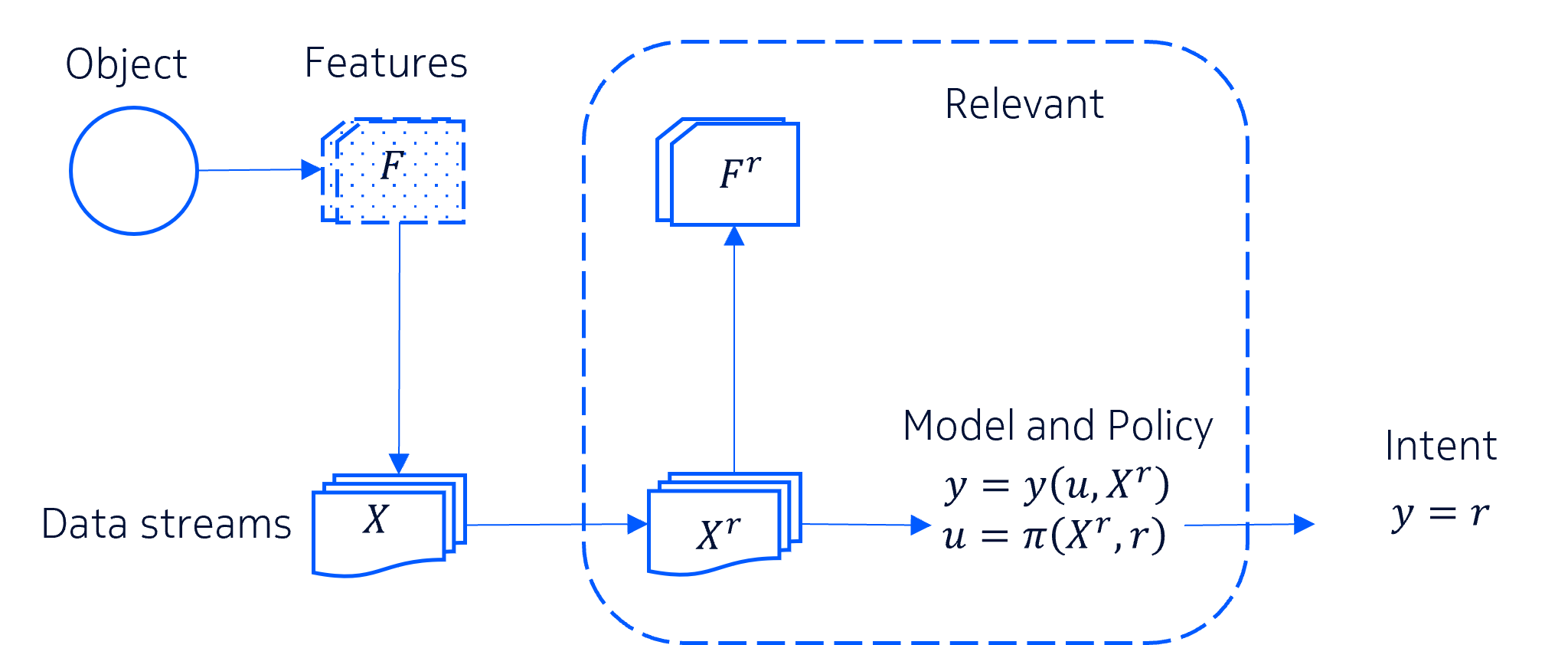}
\caption{Problem formulation. The agent observes the data stream $X$. It needs to identify the relevant ones $X^r$ and define the relevant features $F^r$. It needs to learn the model $y = y(u, X^r)$ and policy $u=\pi(X^r, r)$  to satisfy the intent $y=r$.}
\label{fig:problem}
\end{figure}

The agent receives many data streams from its environment. The data streams are in the form of a vector $X = \{x_0, x_1, x_2, \ldots\}$. They are generated by the features $F=\{f_0, f_1, f_2, \ldots\}$ of objects in the environment. The agent isn't given the features and needs to define them from data streams. Not all features are relevant to the problem. Consequently, not all the data streams are. The set of relevant data streams is $X^r$. The agent needs to identify $X^r$, define the relevant features $F^r$, build a model $y = y(u, X^r)$ from these data streams and find the policies $u=\pi(X^r, r)$ such that $y=r$. In this special case, there is a bijection mapping between data streams and features. Figure \ref{fig:problem} provides a diagram for the problem. 

\begin{figure}[t]
\centering
\includegraphics[width=0.6\columnwidth]{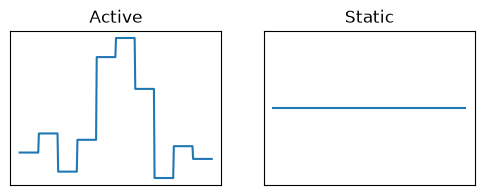}
\caption{Context. Combinations of active and static data streams form contexts. $c = (X^a, X^r \setminus X^a)$, $X^a = (x^a_0 \in V^c_0, x^a_1 \in V^c_1, \ldots)$, and $X^r \setminus X^a = (x^{a-}_0 = v^c_0, x^{a-}_1 = v^c_1, \ldots)$}
\label{fig:context}
\end{figure}

Different combinations of active and static features and their values form different contexts. As shown in Figure \ref{fig:context}, a context is $c = (X^a, X^r \setminus X^a)$ where $X^a = (x^a_0 \in V^c_0, x^a_1 \in V^c_1, \ldots)$ and contains all the active data streams in the context, the set $V^c_i$ contains all possible values of $x^a_i$ in the context, $X^r \setminus X^a = (x^{a-}_0 = v^c_0, x^{a-}_1 = v^c_1, \ldots)$, and $v^c_i$ is the value for $x^{a-}_i$ in the context. After the features are defined, a context becomes $c = (F^a, F^r \setminus F^a)$. The set of active features in the initial context is $F^a_0$. In this special case, the features in $F^r \setminus F^a_0$ stay constant in each context. Their values can change from one context to another. The agent experiences different contexts $c_0, c_1, c_2, \ldots$ over time. The set of contexts it has experiences so far is $C_n = \{c_0, c_1, \ldots, c_n\}$. From the experience $C_n$, the agent needs to identify the relevant data streams $X^r_n \subseteq X^r$ , define the relevant features $F^r_n \subseteq F^r$, build the model for the intent $y = y_n(u, X^r_n)$ and find the policy $u = \pi_n(X^r_n, r)$ to satisfy the intent $y=r$.

\section{The Example} \label{sec:example}
\begin{figure}[t]
\centering
\includegraphics[width=0.6\columnwidth]{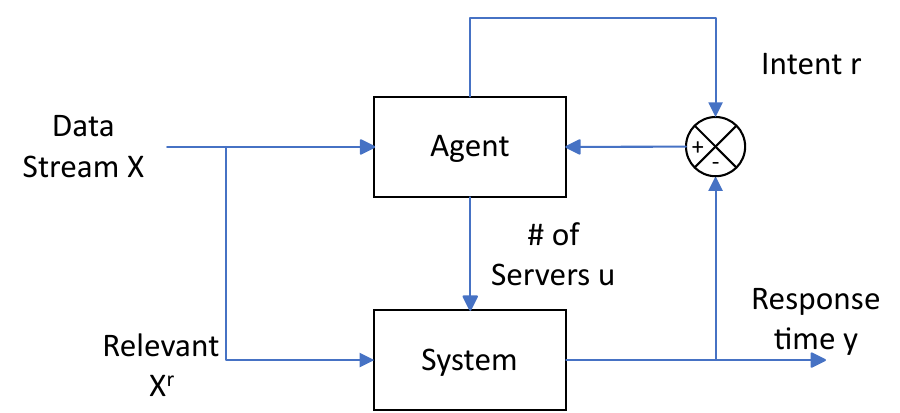}
\caption{The Example. The agent decides the number of servers $u$ to control the response time $y$ in a system. It has an intent to satisfy the requirement of the response time $y=r$. It observes a data stream $X$. It needs to identify the relevant ones $X^r$, define the relevant features $F^r$, and learn the model and policy.}
\label{fig:example}
\end{figure}

I use a stylized example in computing to illustrate the continual learning problem and method. The example is illustrated in Figure \ref{fig:example}. 

An agent manages a pool of servers and has the intent to meet the requirement $r$ for the response time $y$ with the minimal number of servers $u$. It receives data streams from the environment. The data streams are in the form of a vector $X = \{x_0, x_1, x_2, x_3\}$. The agent isn't given the features and needs to define them from data streams. The set of relevant features is $F^r = \{f_0, f_1\}$ where $f_0$ is the arrival rate of packets in a workload and generates $x_0$, and $f_1$ is packet sizes and generates $x_1$. The data streams $x_2$ and $x_3$ are generated by features irrelevant for the intent. The set of active features in the initial context is $F^a_0=\{f_0\}$ and the set of static features is $F^r \setminus F^a_0 = \{f_1\}$. The agent needs to define the relevant features. After the agent receives a sample of data streams, it needs to decide the number of servers $u$, execute the decision and observe the outcome of the response time $y$. It needs to build the model for the intent and find the policy.

\begin{figure}[t]
\centering
\includegraphics[width=0.8\columnwidth]{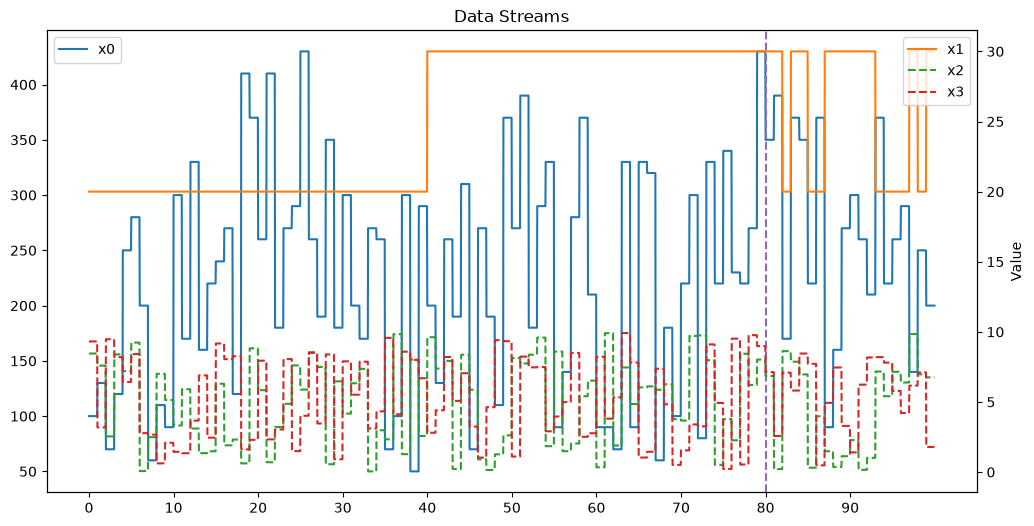}
\caption{Data Stream $X = (x_0, x_1, x_2, x_3)$. Data stream $x_0$ is generated by the arrival rate of packets. Data stream $x_1$ is generated by the packet size. Data streams $x_2$ and $x_3$ are irrelevant. The initial context $c_0$ is from time index 0 to 39. The new context $c_1$ is from time index 40 to 79. Time indices from 80 to 99 have a mixture of contexts $c_0$ and $c_1$ for testing.}
\label{fig:data_stream}
\end{figure}

The agent experiences 2 contexts: an initial context $c_0 = (x_0 \in V_0, x_1 = v^0_1)$, and a new context $c_1 = (x_0 \in V_0, x_1 = v^1_1) $ where $x_1$ takes a new value $v^1_1 \neq v^0_1$. The stream $x_1$ becomes active when it changes value from context $c_0$ to $c_1$. In other words, it is active in the experience $C_1 = \{c_0, c_1\}$. The agent is not told when the context changes. It needs to detect the new context. I show the data streams over time in Figure \ref{fig:data_stream}. The initial context $c_0$ is from time index 0 to 39. The new context $c_1$ is from time index 40 to 79. The time indices from 80 to 99 have a combination of context $c_0$ and $c_1$ for testing. 

I use the detection of context $c_1$ and the learning in $c_1$ to showcase the method. The core of the method is to find the outliers in the data streams when the intent outcome has outliers. Before the context $c_1$ arises, through learning in the initial context $c_0$, the agent has an existing set of relevant data $X^r_0 = \{x_0\}$, an existing set of relevant feature $F^r_0 = \{f_0\}$, an existing model $y = y_0(u, X^r_0)$ and an existing policy $u = \pi_0(X^r_0, r)$. It uses Bayesian Optimization to build the model and find the policy.

\begin{figure}[t]
\centering
\includegraphics[width=0.8\columnwidth]{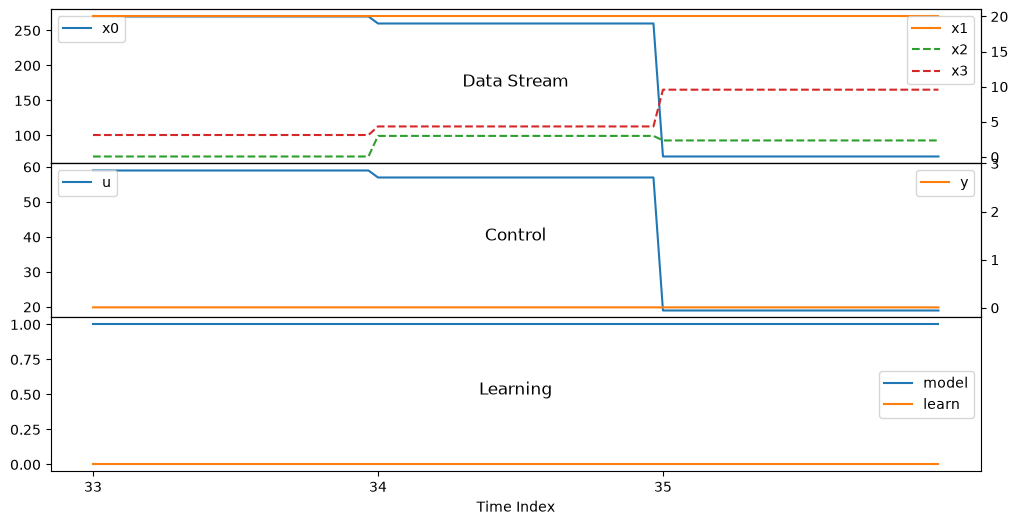}
\caption{Use the policy $\pi_0$ in 3 time indices. The subplot ``Data Stream'' shows the data streams. The subplot ``Control'' shows the action $u$ and the response time $y$ achieved. The subplot ``Learning'' shows the model the agent assigns the sample to and whether the agent performs learning. The model $y_0$ is labeled as 1 and $y_1$ is labeled as 2.}
\label{fig:existing_model}
\end{figure}

The policy $\pi_0$ has been working well for the agent in context $c_0$. When the agent observes a new sample of the relevant data stream $x_0$, it gets a number of servers $u$ from the policy $\pi_0$. It executes the action and observes that the response time $y$ complies to the requirement $r$. More specifically, the agent collects posterior variances during Bayesian Optimization. When a new sample of $x_0$ comes, if the prior variance of the model $y_0$ at $(u = \pi_0(X^r_0, r), X^r_0)$ is within 2 standard deviations of the samples of posterior variances collected, the agent considers that the model $y_0$ has good certainty and it can apply the action $u$ and expect the response time $y = r$. It confirms its expectation after it executes the action. Figure \ref{fig:existing_model} illustrates 3 time indices where the agent uses the policy $\pi_0$ and confirms the intent is satisfied. The subplot ``Data Stream'' shows the data streams. The subplot ``Control'' shows the action $u$ from the policy $\pi_0$ and the response time $y$ achieved. The subplot ``Learning'' shows that the agent uses the model $y_0$ and the agent doesn't performs learning.

\begin{figure}[t]
\centering
\includegraphics[width=0.8\columnwidth]{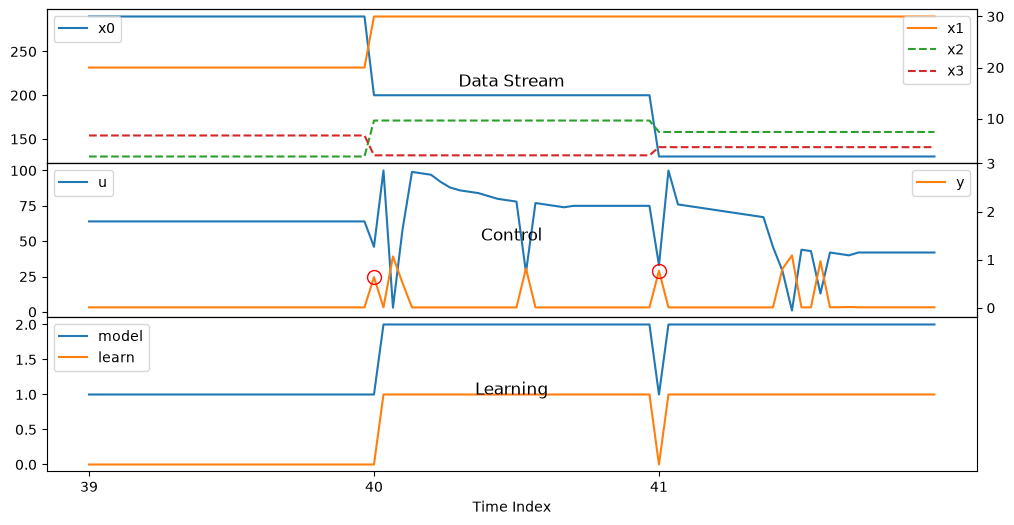}
\caption{Detect new context. The new context arises at the start of time index 40. At first, the agent applies the policy $\pi_0$. It observes an outlier $y \neq r$ in the intent outcome, marked by the red circle. It detects a new context from this outlier. It steps away from the model $y_0$ and starts learning a new model and policy. Towards the end of time index 40, it satisfies the intent again. The same operation repeats in time index 41.}
\label{fig:detect_new_context}
\end{figure}

Then the context $c_1$ arises. The agent is not told when the context changes. But it observes that its policy stops working suddenly: the response time $y$ violates the intent $r$ greatly after it executes the number of servers $u$ from the policy $\pi_0$. The policy should have worked. But it didn't. The agent detects the change of context from the discrepancy between its expectation and observation, i.e. an outlier in the intent outcome. It once again uses Bayesian Optimization and learns a new model $y = y^+_0(u, X^r_0)$ and the associated policy $u = \pi^+_0 (X^r_0, r)$ that makes the response time $y$ meets the requirement $r$ again. Figure \ref{fig:detect_new_context} illustrates the detection and the learning. The new context arises at the start of time index 40. At first, the agent applies the policy $\pi_0$. It observes an outlier $y \neq r$ in the intent outcome as marked by the red circle. It detects a new context from this outlier. It starts learning a new model and policy. Towards the end of time index 40, it satisfies the intent again. The same operation repeats in time index 41.

\begin{figure}[t]
\centering
\includegraphics[width=0.8\columnwidth]{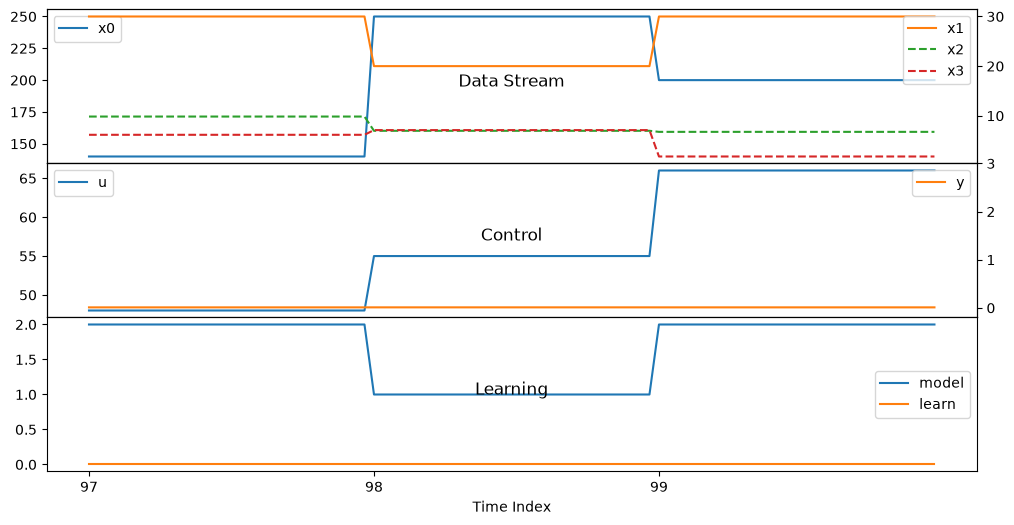}
\caption{Use updated model in 3 time indices. The values of $x_1$ are different. The agent selects the model and applies the policy accordingly and observes that the intent is satisfied.}
\label{fig:updated_model}
\end{figure}

After the agent encounters the new context $m$ times, it seeks to find a new relevant feature and distinguishes the contexts. For each data stream in $X \setminus X^r_0 = \{x_1, x_2, x_3\}$, it compares each sample in the new context to the statistics in context $c_0$ and computes the ratio of outliers. More specifically, it computes z-scores for the samples based on the mean and standard deviation in context $c_0$; if a z-score is above 3, it observes an outlier. Since the intent outcomes in context $c_1$ are all outliers for context $c_0$, i.e. $y\neq r | \pi_0$, this operation correlates the outliers in the intent outcomes to those in the data streams. The agent finds that $x_1$ has a ratio of (nearly) 1. As shown in Figure \ref{fig:detect_new_context}, the outliers of the intent outcome coincide with the outliers of the data stream $x_1$. The agent identifies this stream as the feature that distinguishes contexts. It defines it as feature $f_1$, adds it to the set of relevant features and expands the inputs of the model and the policy by conditioning their selection on the value of $x_1$. 
The agent does the following updates.
\begin{eqnarray}
F^r_1 &=& F^r_0 \cup \{f_1\}\\
X^r_1 &=&  X^r_0 \cup \{x_1\}\\
y = y_1(u, X^r_1) &=& 
\begin{cases}
y_0(u, X^r_0), & \text{ if } x_1 = v^0_1 \\
y^+_0(u, X^r_0), & \text{ if } x_1 = v^1_1
\end{cases}\\
u = \pi_1(X^r_1, r) &=& 
\begin{cases}
\pi_0(X^r_0, r), & \text{ if } x_1 = v^0_1 \\
\pi^+_0(X^r_0, r), & \text{ if } x_1 = v^1_1
\end{cases}
\end{eqnarray}
From this point on, the agent applies the policy $\pi_1$ which conditions the selection of the actions on the value of $x_1$. Figure \ref{fig:updated_model} illustrates the agent uses the updated model. The 3 time indices have different values for $x_1$. The agent selects the model and applies the policy accordingly and observes that the intent is satisfied. 

The parameter in the examples are as follows. The requirement for the response time $r = 0.011$. The values for the active feature $x_0$ are randomly drawn from $V_0 = \{50, 60, \ldots, 450\}$. The values for the static feature $x_1$ are $v^0_1 = 20$ and $v^1_1=30$. The service rate $\mu$ of the servers is $100$. The response time $y$ is computed approximately using a M/M/c queue, $y = \frac{C(c, \lambda/\mu)}{c\mu - \lambda} + \frac{1}{\mu}$ where $C$ is the Erlang'c C formula, $c$ is the number of servers ($u$ in my notation), $\lambda = x_0 \times x_1$. The range for the number of servers $u$ is $[1, 100]$.  The agent accumulates $m=30$ occurrences to identify the new feature.

\section{The Method} \label{sec:method}

\begin{figure}[t]
\centering
\includegraphics[width=\columnwidth]{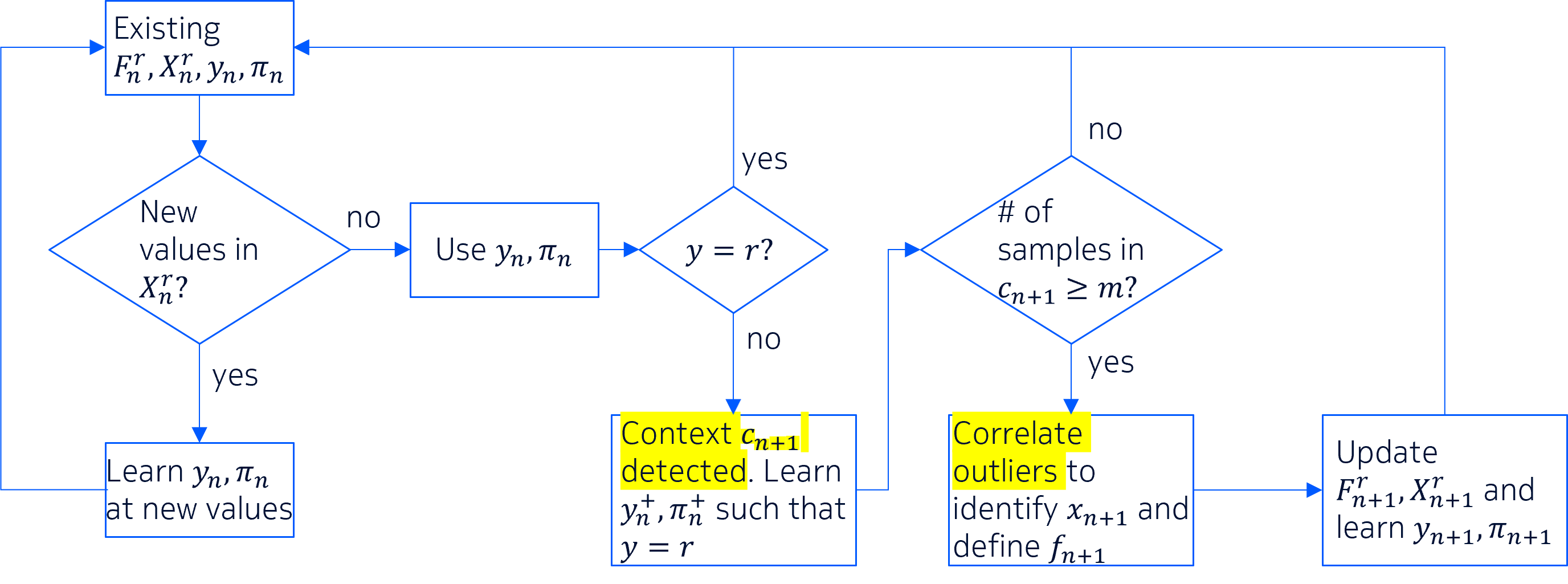} 
\caption{Outline of Homeostatic Continual Learning method. The agent detects new context from the outliers where $y\neq r$ and identifies new relevant data stream $x_{n+1}$ and new feature $f_{n+1}$ through correlating the outliers in the data streams $X \setminus X^r_n$ with the outliers in the intent outcomes.}
\label{fig:method}
\end{figure}

I call my method ``Homeostatic Continual Learning''. The homeostatic mechanism in biology regulates changes and maintains steady internal conditions of an life form. The agent in my work keeps the outcome $y$ of the intent to the reference $r$. I use the word ``homeostatic'' to highlight the analogy with the homeostatic mechanism and to indicate that my method imitates the one animals use. The method is outlined in Figure \ref{fig:method}. Its core is to find the outliers in the data streams when the intent outcome has outliers. 

The agent starts with an existing model $y_n$ and the associated policy $\pi_n$ that satisfies an intent with a practical level of certainty $\alpha$ in existing experience $C_n$. The policy $\pi_n$ is the optimal action for given criteria. It has a set of relevant data streams $X^r_n$ and a set of relevant features $F^r_n$. If there is no new context, when the agent receives a sample of the data, it applies the policy $\pi_n$ and gets an action. It executes the action and observes that the intent outcome $y$ complies to the reference $r$. 

When a new context arises, it can be new values for a feature in the relevant feature set $F^r_n$ or one in the set $F^r \setminus F^r_n$. If new values appear for a feature in $F^r_n$, the model $y_n$ has low certainty at the new values as the agent hasn't experienced the values before. The agent observes this low certainty, doesn't use the existing policy and learns the model $y_n$ and the policy $\pi_n$ at the new values. 

If new values appear for a feature in $F^r \setminus F^r_n$, the agent observes that its policy stops working suddenly: the intent outcome $y$ violates the reference $r$ when it applies the action from the policy $\pi_n$. The agent detects a new context $c_{n+1}$ from the outlier in the intent outcome. It learns a new model $y = y^+_n(u, X^r_n)$ and the associated policy $u = \pi^+_n (X^r_n, r)$ that makes the response time $y$ meets the requirement $r$ again. 

After the agent encounters the new context $m$ times, it seeks to find the new feature that distinguishes the contexts. For each data stream in $X \setminus X^r_n$, it compares each sample in context $c_{n+1}$ to the statistics in the existing experience $C_n$ and computes the ratio of outliers. Since the intent outcomes in context $c_{n+1}$ are all outliers in the existing experience $C_n$, i.e. $y \neq r | \pi_n$, this operation correlates the outliers in intent outcomes to those in the data streams. The agent selects the stream with the highest ratio, labels it as $x^r_{n+1}$ and defines a feature $f^r_{n+1}$ out of it. It expands the set of relevant features $F^r_{n+1} = F^r_n \cup \{f^r_{n+1}\}$ and the inputs $X^r_{n+1} = X^r_n \cup \{x^r_{n+1}\}$ of the model and the policy. It combines its experiences and learns a new model $y = y_{n+1}(u, X^r_{n+1})$ and the associated policy $u = \pi_{n+1}(X^r_{n+1}, r)$. It does the following updates. 
\begin{eqnarray}
F^r_{n+1} &=& F^r_n \cup \{f^r_{n+1}\} \\
X^r_{n+1} &=& X^r_n \cup \{x^r_{n+1}\} \\
y &=& y_{n+1}(u, X^r_{n+1})\\
u &=& \pi_{n+1}(X^r_{n+1}, r)
\end{eqnarray}
One possibility for $y = y_{n+1}(u, X^r_{n+1})$ and $u = \pi_{n+1}(X^r_{n+1}, r)$ is
\begin{eqnarray}
y = y_{n+1}(u, X^r_{n+1}) =
\begin{cases}
y_n(u, X^r_n), & \text{ if } x^r_{n+1} = v^n_{n+1} \\
y^+_n(u, X^r_{n}), & \text{ if } x^r_{n+1} = v^{n+1}_{n+1}
\end{cases}\\
u = \pi_{n+1}(X^r_{n+1}, r) = 
\begin{cases}
\pi_n(X^r_n, r), & \text{ if } x^r_{n+1} = v^n_{n+1}  \\
\pi^+_n(X^r_n, r), & \text{ if }  x^r_{n+1} = v^{n+1}_{n+1}
\end{cases}
\end{eqnarray}
From this point on, it applies the model $y_{n+1}$ and policy $\pi_{n+1}$. Next time a new context arises, the agent repeats the same procedure. 

The intent and a notion of certainty are critical for the method. The intent dictates the relevant features and guarantees the performance in different contexts. The notion of certainty enables the detection of new contexts. 

I recall that this method is for the special case where 
\vspace{\baselineskip}
\begin{enumerate}[label=\textnormal{(\Roman*)}]
\item There is a bijection mapping between data streams and features. 
\item A static feature stays constant in all the contexts. It only becomes active by changing values from one context to another.
\item From one context to another, only 1 static feature changes values. The active features have the same set of values.
\end{enumerate}
\vspace{\baselineskip}

There are 2 cases Assumption \RomanNumeralCaps{1} is violated. The first case is that one feature generates multiple data streams, $f_j \equiv X^j = \{x^j_0, x^j_1, \ldots\}$. The agent identifies the data streams through the same procedure and defines a new feature out of the vector of the identified data streams. The second case is that multiple data streams $\{x_0, x_1, \ldots\}$ are jointly generated by multiple features $\{f_0, f_1, \ldots\}$ and no stream corresponds to individual features. One example is the picture of an object is generated by the object and the lighting condition. The agent doesn't have separate data streams for active and static features. One possible remedy is to for the agent to extract the feature as statistics of the data streams and uses $F^r$ instead of $X^r$ in the model and policy.

When Assumption \RomanNumeralCaps{2} is violated, a static feature may become active in another context. The agent handles this case through the same procedure. No modification is needed.

For Assumption \RomanNumeralCaps{3}, there are also 2 cases it is violated. One case is for multiple static features to change values at the same time. The agent treats this situation as in the first way Assumption \RomanNumeralCaps{1} is violated. It defines a new feature out of the vector of the identified data streams $f_j \equiv X^j = \{x^j_0, x^j_1, \ldots\}$. If later on some of these static features change values while the others don't and generate $X^j = (X^j_a, X^j \setminus X^j_a)$ where the data streams in $X^j_a$ change value while those in $X^j \setminus X^j_a$ don't, the agent splits the data streams and define multiple new features. It removes $f_j$ and adds new features $f^a_{j} \equiv X^{j}_a$ and $f^{a-}_{j} \equiv X^j \setminus X^j_a$. The other case is that active features and static features have new values at the same time. The agent learns the model $y_n$ and policy $\pi_n$ at the new values of the active features and thus introduces errors in its model and policy. It can correct the errors when it identifies the static features from other contexts.

\section{The Findings} \label{sec:findings}

The method solves the catastrophic forgetting problem. The root cause of the catastrophic forgetting is the model uncertainty. When the new context arises and a relevant feature becomes active, the existing model is no longer complete. A single input generates 2 different outputs. Reinforcement Learning migrates to the output in the new context through value updates and forgets the output in the previous contexts. With Homeostatic Continual Learning, the agent identifies the new relevant feature. By incorporating the feature into its model, it resolves the model uncertainty and restore the uniqueness of the output. This resolves the catastrophic forgetting problem. The method produces policies adapted to contexts and maintains the satisfaction of the intent.

\begin{figure}[t]
\centering
\includegraphics[width=0.6\columnwidth]{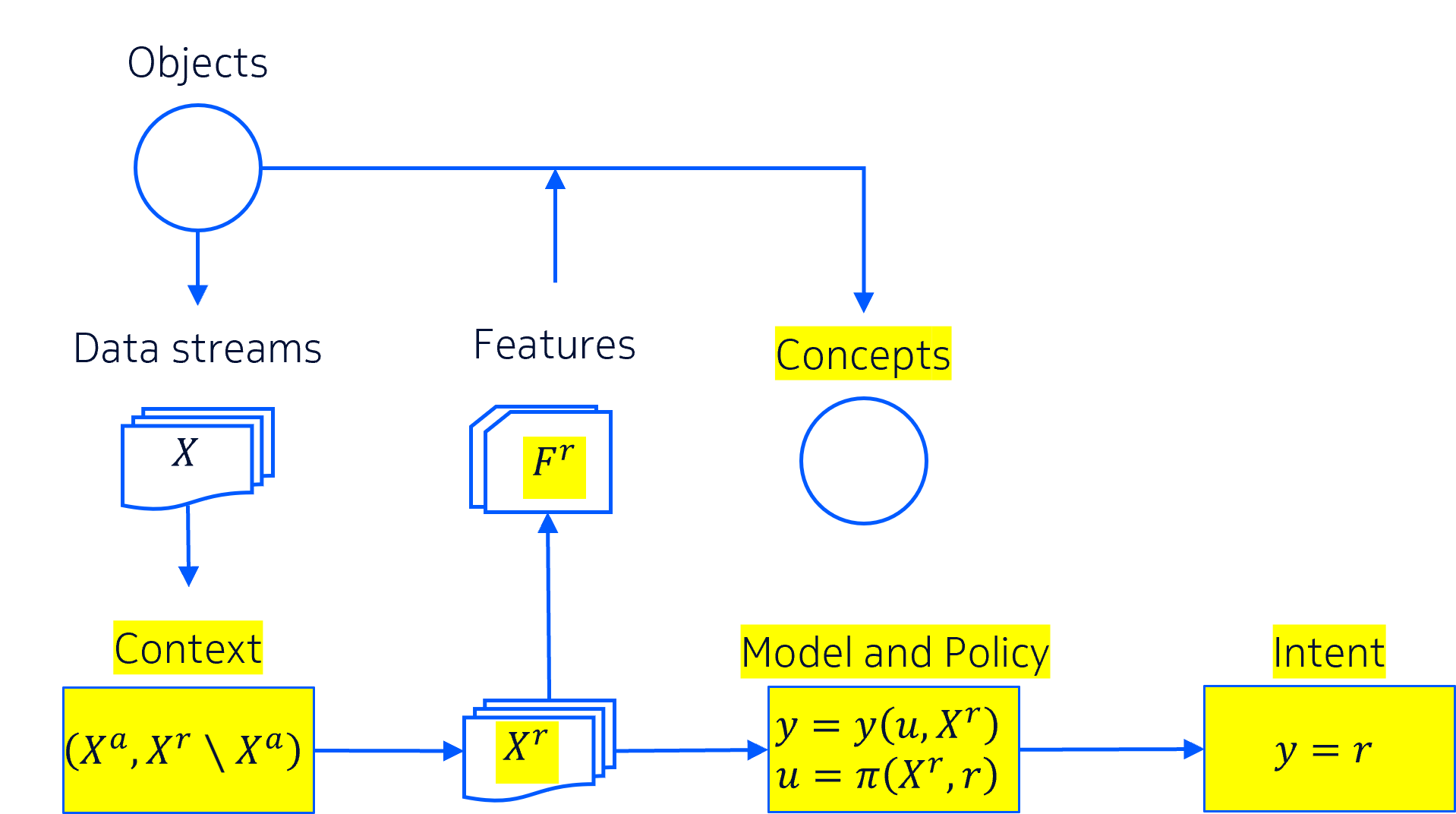}
\caption{Learning process. From the contexts, the agent identifies the relevant data streams $X^r$, defines the features $F^r$ out of $X^r$ and learns models $y(u, X^r)$ and policies $\pi(X^r, r)$ to satisfy its intent $y=r$. Through the features $F^r$, it abstracts individual objects into comparable instances of concepts.}
\label{fig:learning_process}
\end{figure}

\begin{figure}[t]
\centering
\includegraphics[width=0.6\columnwidth]{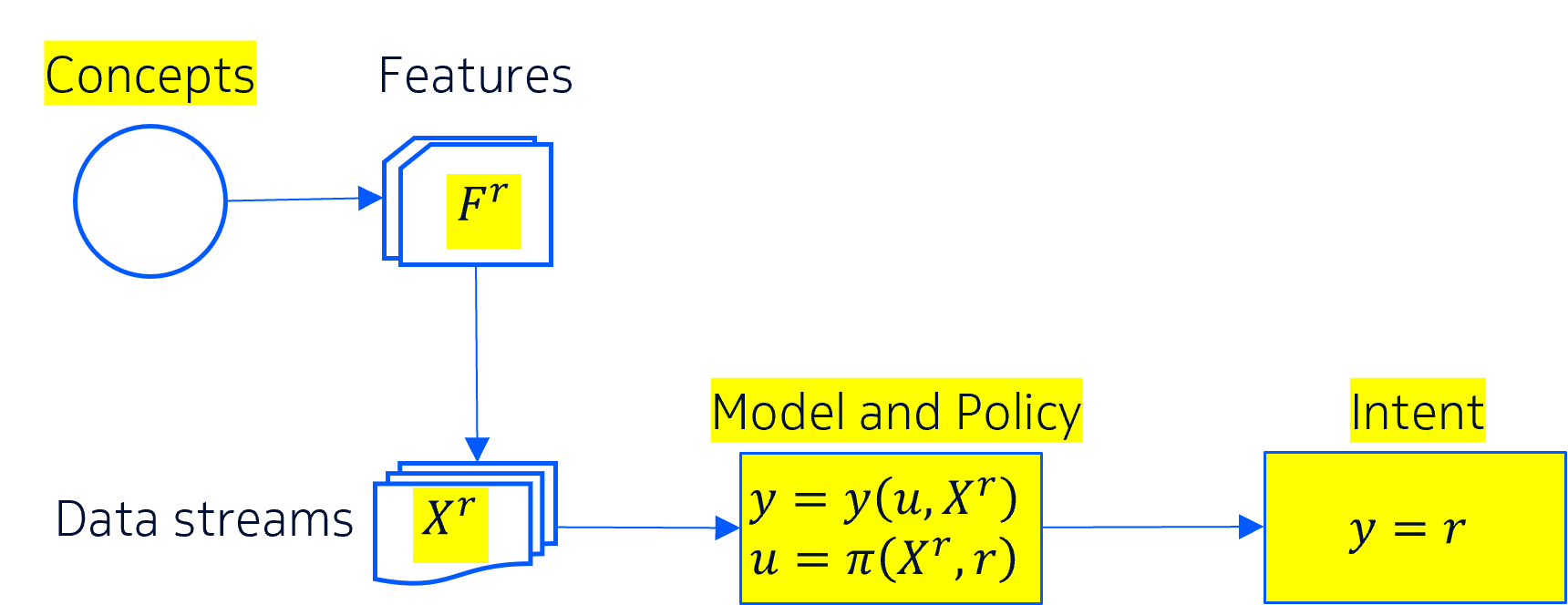}
\caption{Learned knowledge. Concepts have features $F^r$. Features $F^r$ generate data streams $X^r$. The model $y(u, X^r)$ and policy $\pi(X^r, r)$ are learned based on the data streams $X^r$ to satisfy the intent $y=r$.}
\label{fig:learned_knowledge}
\end{figure}

The method enables the abstraction of objects in the environment. Figure \ref{fig:learning_process} outlines the learning process. From the contexts, the agent identifies the relevant data streams $X^r$, defines the features $F^r$ out of $X^r$ and learns models $y(u, X^r)$ and policies $\pi(X^r, r)$ to satisfy its intent $y=r$. Through the features $F^r$, it abstracts individual objects into comparable instances of concepts. Figure \ref{fig:learned_knowledge} outlines the learned knowledge. Concepts have features $F^r$. Features $F^r$ generate data streams $X^r$. The model $y(u, X^r)$ and policy $\pi(X^r, r)$ are learned based on the data streams $X^r$ to satisfy the intent $y=r$. This abstraction from individual objects to concepts greatly reduces the demand of computing resources on the agent. Without the abstraction, the agent needs to map specific objects to intents; with it, the agent only needs to map features to intents. For instance, if the agent needs to ``select 4 specific apples from 100 apples'', it needs to select among $\binom{100}{4}$ choices; but if the agent needs to ``get 1 kilogram of big red apples'', it can randomly choose some big red apples that weight 1 kilogram together. The reduction of complexity is enormous. 

Moreover, the agent can combine the features it has learned to better adapt to new contexts. Table \ref{tb:combine_features} gives an example of combining features. When the agent has driven on ``dry and flat'' roads, ``dry and steep'' roads and ``icy and flat'' roads, it has some idea how it should drive on ``steep and icy'' roads before its first trial.

\begin{table}[t]
\centering
\begin{tabular}{l|cc}
    \hline
     & Dry road & Icy road \\
    \hline
    Flat road & X & X \\
    Steep road & X & (?)\\
    \hline
\end{tabular}
\caption{An example of combining features to better adapt to new contexts. When the agent has driven on ``dry and flat'' roads, ``dry and steep'' roads and ``icy and flat'' roads, it has some idea how it should drive on ``steep and icy'' roads before its first trial.}
\label{tb:combine_features}
\end{table}

\begin{figure}[t]
\centering
\includegraphics[width=0.5\columnwidth]{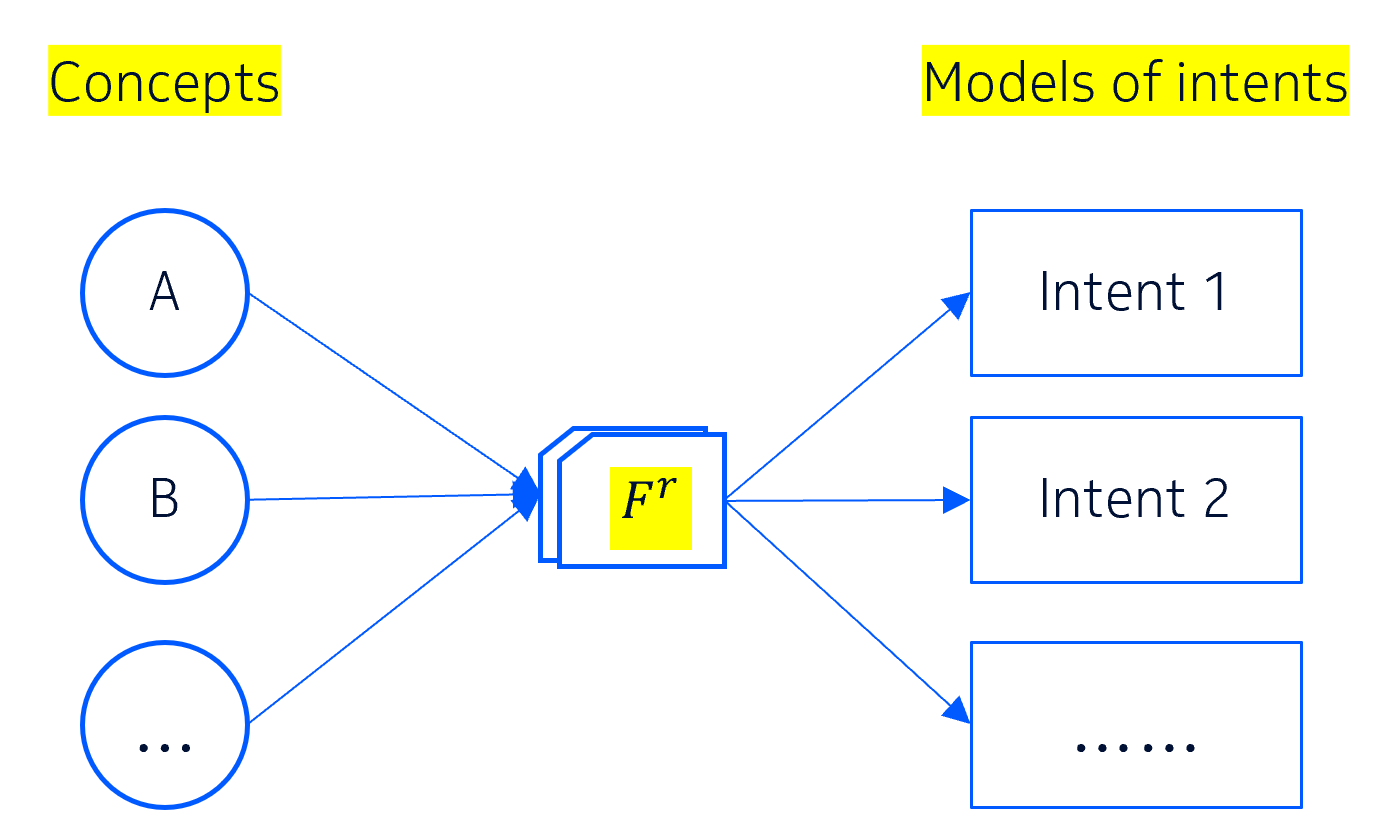}
\caption{World Model. It has a structure similar to a transformer. The symbolic features are derived from sub-symbolic data streams. Concepts are linked to intents through features. }
\label{fig:world_model}
\end{figure}

With the method, the agent builds a world model by mapping features to intents and abstracting objects into concepts when it pursuits many intents concurrently. The essence of building a world model is to factorize the world into features. Figure \ref{fig:world_model} illustrates the structure of the world model. It is similar to a transformer \cite{vaswani2017attention}. The symbolic features are derived from sub-symbolic data streams. Concepts are linked to intents through features. A ``Feature'' node serves as a switch for selecting concepts for intents. This is similar to the queries in transformers. The nodes in the world model, including features, concepts and intent models, and the links between them are all semantic. The world model is explainable by nature. Each time the agent updates its model and policy, it can explain the reason for the update and the streams / features it identifies.

The method has many other merits. It doesn't require labeled data. It naturally adapts to the time scale of context change. The frequency of the context change drives the frequency the agent updates its model and policy. The update is done on selective data and features and is memory efficient. The method is applicable at many levels of problem complexities.

\section{The Works} \label{sec:works}

\begin{figure}[t]
\centering
\includegraphics[width=0.6\columnwidth]{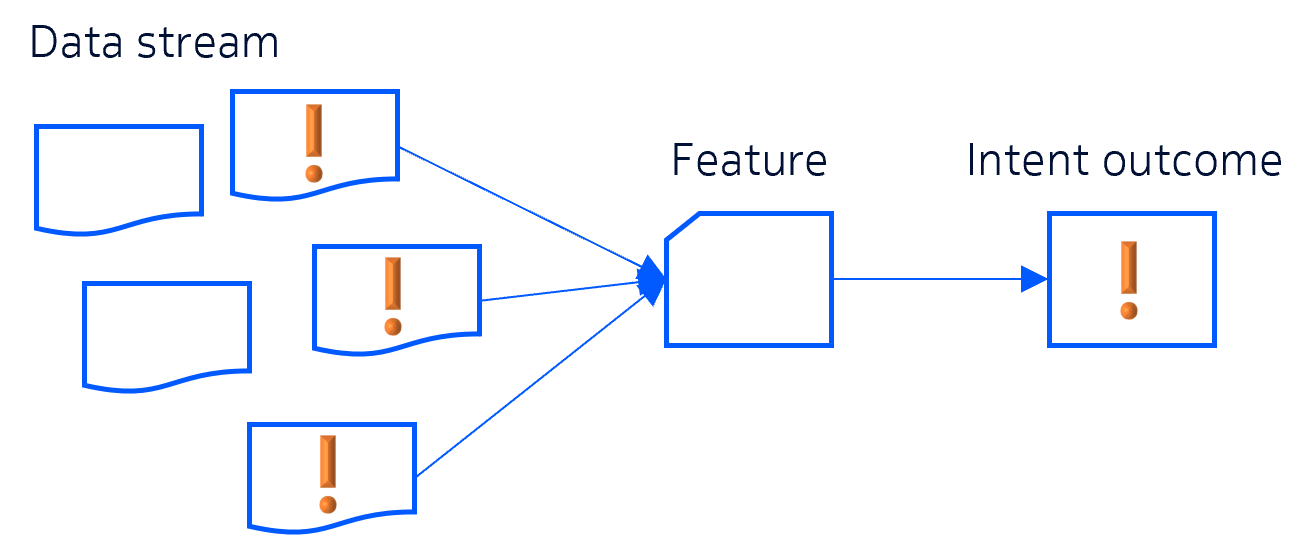}
\caption{Use Spike-Timing-Dependent Plasticity.  When the intent outcome has an outlier, the agent looks for outliers in the data streams. It links all the data streams that have outliers to a node and uses the node to represent the feature. It links the feature to the intent. }
\label{fig:STDP}
\end{figure}

We need to undertake many works to render the method practical. We need to implement this method with neural network. The Spike-Timing-Dependent Plasticity (STDP) \cite{caporale2008spike} fit naturally with the method. In STDP, ``neurons that fire together wire together''. Figure \ref{fig:STDP} shows that how STDP can be used for the method. When the intent outcome has an outlier, the agent looks for outliers in the data streams. It links all the data streams that have outliers to a node and uses the node to represent the feature. It links the feature to the intent. This operation is energy efficient as only a small number of nodes are involved. As the agent gradually completes its world model, the nodes are connected gradually and form a network that appears deep. If we trace the neuron firing sequences, we understand the inner thinking of the agent as the neurons are semantic. 

We also need to 
\begin{itemize}
\item Implement the method at scale. The agent may apply the method on multiple intents concurrently. If the intents are interconnected, the agent needs methods to update its world model coherently and make trade-offs between the actions in the intents. 
\item Develop an approach for the agent to extract and transfer knowledge of the effect of a feature. In the world model, the agent maps a feature to multiple intents. When the agent learns the effect of a feature from an intent, it needs a method to extract that information and transfer it to another intent. 
\item Develop an approach for the agent to induce recursively, abstract the concepts further, and create a hierarchy of concepts. For example, an approach enables the agent to abstract ``apple'' into ``fruit'' and further into ``food''. 
\item Develop an approach for the agent to correct the errors it makes. For example, the agent needs a way to correct the error it makes in the second case Assumption \RomanNumeralCaps{3} is violated. 
\item Develop a version of the method where an agent teaches another agent about the effect of a feature after its own learning. What data should the teacher agent give to the student agent? What metrics can the teacher use to assess the learning of the student?
\end{itemize}

\section{The Connections} \label{sec:connections}

The method connects to many other fields in Artificial Intelligence. It connects to Causal Discovery. The new feature in the context is candidate for the cause. The agent then intervenes on the feature and tests whether it changes the intent outcome. If it does, the agent confirms causality. If not, a hidden confounder exists and the feature represents the effect of the confounder; the agent needs to search for additional information. In Pavlov's experiment, the confounder is Pavlov's action of conducting the experiment. The confounder produces both the bell and the food. If the agent intervenes on the bell and observes that food doesn't appear, it detects a hidden confounder. It needs to search further for the real cause.

It connects to optimization. It resolves the model uncertainty issue in optimization. The method solves optimization as a series of constraint satisfaction problems, as shown in Equation \ref{eq:constraint_satisfaction}. The agent has an intent to continuously improve performance. It starts with a certain intent reference $r$. Once its policy achieves certainty in obtaining $r$ in all the existing contexts, it increases the value of $r$. Over many iterations, the agent finds the maximum intent reference it can satisfy. The agent uses the same method to optimize its own learning processes. It has intents to effectively and efficiently use its computing resources. It uses the method to learn a model for using its computing resources and the associated policies and optimizes over the learning rate, the sensitivity in detecting new contexts, the intensity and duration for learning new model and policy, etc. to satisfy these intents.

\begin{equation}
\begin{aligned} 
&& min_x f(x) \\ 
s.t. && g_i(x) \leq 0 \\ 
&& h_j(x) =0
\end{aligned}
\quad \Longrightarrow \quad
\begin{aligned} 
&& f(x)=r \\ 
&& g_i(x) \leq 0 \\ 
&& h_j(x) =0
\end{aligned}
\label{eq:constraint_satisfaction}
\end{equation}

The method also connects to 
\begin{itemize}
\item Reinforcement Learning. The agent learns to define the states itself with the method. The agent keeps the intent outcomes $y$ to the reference $r$. This collapses the state space for upcoming states and simplifies the transition chains in Markov Decision Processes.
\item Active Learning. The agent can proactively create samples in the new context instead of waiting passively for them to happen. If the agent can't find a data stream that distinguishes the contexts in existing data streams, it needs to search for new data streams.
\item One-Shot / Few Shot Learning. I propose that one-shot or few-shot learning is essentially creating and testing new combinations of known features and their effect. \cite{shi2026programmingparadigmspatiotemporalcomposability} provides a potential technique.
\item Cognitive Architecture. We need to design the intents the agent has. This links to the Principle-Agent problem in economy \cite{Grossman1992} where agents generally do not maximize what the principal wants; they maximize what the principal rewards. The intents need to be adequate for the agents to provide the desired outcomes. We need to design the starting point of an agent. What should the agent know or be capable of doing at the beginning? We need to design the maximum cognitive capacity of an agent which I think is tied to the amount of neurons it has. 
\item Federated Learning. We need to decide the number and types of agents for managing large systems such as networks: whether it should be a single agent with an enormous amount of computing resources or a group of agents with a moderate amount. We need to design the collaboration protocols between the agents. Alternatively, we need to develop approaches for the agents to learn the protocols themselves.
\end{itemize}

\section{Conclusion} \label{sec:implications}

When an agent operates in an open world, it needs to learns what affects its intents from its experiences. It does so with Homeostatic Continual Learning. The core of the method is to find the features that have outliers when the intent outcome has outlier. The agent gradually completes its model and policy and performs well in more and more contexts through the method. As a result of the learning, the agent builds a transformer-like world model where it factorizes the objects in the world into features, abstract objects into comparable instances of concepts and map concepts to intents through features. The world model is composed of semantic nodes and explainable by nature. The dog in Pavlov's experiment has learned many contexts in which it gets food before the experiment. When it gets food in the experiment, this becomes an outlier in its intent outcome. It instinctively searches for the features that have outliers at the same time. It finds the sound of the bell. It links the sound of the bell to food and updates its world model with one more feature relevant for food.

We can explain many other psychological phenomena through this method. When babies are born, they exhibit primary narcissism because they don't know yet what external features to condition their actions on. Emotions are perceived patterns of cognitive operations in Homeostatic Continual Learning \cite{jin2026emotionsrecognizedpatternscognitive} which the author refers to as ``Closed Control Loop''. Through implementing and training an autonomous agent, we will have the opportunities to observe a development process similar to human mind and deepen our understanding of ourselves. 

If we simply give the agent Key Performance Indicators (KPIs) as intents, we could run into the Principle-Agent problem \cite{Grossman1992}. What the AI agent does may not meet our real intents and may even harm them. As elaborated in \cite{r.dawkins1976the-selfish-gen}, the root intent in biology is to maximize the chance of the genes being passed down forever. He also coins the word ``meme'' that means an idea, behavior, or style that spreads by means of imitation from person to person within a culture because of the encoding capability of language. I think the root intent in Psychology is to craft one's own memes and maximize the chance of the memes being passed down forever. Should we give an agent a root intent? If so, what should we give? 

Homeostatic Continual Learning connects to many fields in Artificial Intelligence and many works are to be done to release its full power. We need to implement the method with neural network, at scale and with self correction. We need to update Machine Learning and optimization methods in the presence of Homeostatic Continual Learning. We need to revise cognitive architecture to incorporate Homeostatic Continual Learning. We need to learn about human intelligence through observing artificial intelligence. With these works, we can unlock the next phase of artificial intelligence ...... and human intelligence.

\clearpage
\bibliography{../References/gainExperiences}

\end{document}